\pdfoutput=1
\documentclass[11pt]{article}
\usepackage{authblk}
\usepackage{emnlp2021}

\usepackage{times}
\usepackage{latexsym}
\usepackage{enumitem}
\usepackage{threeparttable}

\usepackage[T1]{fontenc}

\usepackage[utf8]{inputenc}

\usepackage{microtype}

\usepackage{amsfonts}
\usepackage{amsmath}
\usepackage{algpseudocode}
\usepackage{algorithm} 
\usepackage{graphicx}
\usepackage{subcaption}
\usepackage{svg}
\usepackage{tikz-cd}
\usepackage{booktabs}
\usepackage{makecell}
\usepackage{multirow}
\usepackage{multicol}

\usetikzlibrary{positioning}

\graphicspath{{images/}}

\title{Artificial Text Detection via Examining the Topology of Attention Maps}

\author{
    ~\textbf{Laida Kushnareva\textsuperscript{1}}\thanks{\ \ Equal contribution.}, 
    ~\textbf{Daniil Cherniavskii\textsuperscript{2}$^*$},
    ~\textbf{Vladislav Mikhailov\textsuperscript{3}$^*$}, \\
    ~\textbf{Ekaterina Artemova\textsuperscript{1,4}}, 
    ~\textbf{Serguei Barannikov\textsuperscript{2,5}},
    ~\textbf{Alexander Bernstein\textsuperscript{2}}, \\
    ~\textbf{Irina Piontkovskaya\textsuperscript{1}}, 
    ~\textbf{Dmitri Piontkovski\textsuperscript{4}},
    ~\textbf{Evgeny Burnaev\textsuperscript{2}} \\ \\
    \textsuperscript{1}Huawei Noah’s Ark lab,
    \textsuperscript{2}Skolkovo Institute of Science and Technology, \\
    \textsuperscript{3}SberDevices,
    \textsuperscript{4}HSE University,
    \textsuperscript{5}CNRS, IMJ
}

\begin{document}

\maketitle

\begin{abstract}
The impressive capabilities of recent generative models to create texts that are challenging to distinguish from the human-written ones can be misused for generating fake news, product reviews, and even abusive content. Despite the prominent performance of existing methods for artificial text detection, they still lack interpretability and robustness towards unseen models. To this end, we propose three novel types of interpretable topological features for this task based on Topological Data Analysis (TDA) which is currently understudied in the field of NLP. We empirically show that the features derived from the BERT model outperform count- and neural-based baselines up to 10\% on three common datasets, and tend to be the most robust towards unseen GPT-style generation models as opposed to existing methods. The probing analysis of the features reveals their sensitivity to the surface and syntactic properties. The results demonstrate that TDA is a promising line with respect to NLP tasks, specifically the ones that incorporate surface and structural information.
\end{abstract}

\section{Introduction}
Recent text generation models (TGMs) based on the transformer architecture \citep{vaswani2017attention} have demonstrated impressive capabilities of creating texts which are very close to human in terms of fluency, coherence, grammatical and factual correctness \citep{keskar2019ctrl,zellers2019defending,yang2019xlnet}. Extensive GPT-style TGMs~\citep{gpt2paper} have achieved outstanding results over a great scope of NLP tasks employing zero-shot, one-shot, and few-shot techniques, even outperforming state-of-the-art fine-tuning approaches \citep{brown2020language}. However, such models can be misused for generating fake news \citep{zellers2019defending,uchendu-etal-2020-authorship}, product reviews \citep{adelani2020generating}, and even extremist and abusive content \citep{mcguffie2020radicalization}. 

Many attempts have been made to develop artificial text detectors \citep{jawahar2020automatic}, ranging from classical ML methods over count-based features \citep{uchendu2019characterizing} to advanced transformer-based models \citep{adelani2020generating} and unsupervised approaches \citep{solaiman2019release}. Despite the prominent performance of these methods across various domains, they still lack interpretability and robustness towards unseen models.

This paper introduces a novel method for artificial text detection based on Topological Data Analysis (TDA) which has been understudied in the field of NLP. The motivation behind this approach relies on the fact that (i) the attention maps generated by the transformer model can be represented as weighted bipartite graphs and thus can be efficiently investigated with TDA, (ii) TDA methods are known to capture well surface and structural patterns in data which, we believe, are crucial to the task. 

The contributions are summarized as follows. (i) To the best of our knowledge, this work is the first attempt to apply TDA methods over the transformer model's attention maps and interpret topological features for the NLP field. (ii) We propose three types of interpretable topological features derived from the attention graphs for the task of artificial text detection. We empirically show that a simple linear classifier trained on the TDA features produced over BERT attentions~\citep{devlin2019bert} outperforms count- and neural-based baselines up to 10\%, and can perform on par with the fully fine-tuned BERT model across three domains: social media, news articles and product reviews. (iii) Testing the robustness towards unseen TGMs, we find that the TDA-based classifiers tend to be more robust as opposed to the existing detectors. (iv) The probing analysis of the features demonstrates their sensitivity to surface and syntactic properties. (v) Finally, we are publicly releasing the code\footnote{\url{https://github.com/danchern97/tda4atd}}, hoping to facilitate the applicability of the TDA methods to other NLP tasks, specifically the ones that incorporate structural information.

\section{Related Work}
\paragraph {Applications of Topological Data Analysis}
TDA has been applied in NLP to study textual structural properties, independent of their surface and semantic peculiarities. These applications include detection of children and adolescent writing \citep{zhu2013persistent}, discourse and entailment in law documents \citep{savle-etal-2019-topological}, and exploring discourse properties of the plot summary to identify the movie genre \citep{doshi2018movie}. \citet{guan2016collapse} apply the topologically motivated transformation of the document's semantic graph to summarize it further. However, these studies neither incorporate neural data representations nor explore the properties of neural language models. 

The research in the emerging scope of TDA applications to neural networks and neural data representations has mainly focused on artificial datasets or common problems in computer vision. The desired topological properties of the data representation can be incorporated into the objective function during the training of a neural network, improving its robustness and performance on the downstream tasks such as human action recognition and image classification \citep{som2020pinet}, image simplification \citep{solomon2021fast}, image segmentation \citep{clough2020topological} or generation \citep{bruel2019topology}. Another line aims to develop the topological criteria of the network's generalization properties \cite{rieck2018neural,corneanu2020computing, naitzat2020topology, barannikov2020topological} or its robustness to adversarial attacks \citep{corneanu2019does}.

\paragraph{Exploring Attention Maps} 
Several studies have shown that attention maps of pre-trained language models (LMs) capture linguistic information. For the sake of space, we will discuss only a few well-known recent works. \citet{clark2019does} attempt to categorize the types of attention patterns observed in the BERT model. In particular, they discover certain attention heads in which prepositions attend to their objects or coreferent mentions attend to their antecedents. Further, they explore the typical behavior of the attention heads and introduce five patterns, e.g. attending to the next token or previous token, which the vast majority of the attention heads follow. 
\citet{htut2019attention} explore the syntactic information encoded in intra-word relation in the attention maps. A maximum spanning tree (MST) is constructed from the computed attention weights and mapped to the corresponding dependency tree for a given sentence. This method achieves a prominent Undirected Unlabeled Attachment Score (UUAS), indicating that the attention graphs indeed can capture the dependency-based relations.
\citet{NEURIPS2019_2c601ad9} explore the importance of the attention heads with respect to a downstream task. They show that a large proportion of the attention heads can be pruned without harming the model downstream performance. Beneficially, the pruned model speeds up at the inference time.
Finally, visualization of the attention maps \cite{exbert} allows introspecting the model's inner workings interactively.

\paragraph{Supervised Artificial Text  Detectors} Several well-established classical ML methods have been applied to the task of artificial text detection combined with topic modeling and linguistic features \citep{manjavacas2017assessing,uchendu2019characterizing,uchendu-etal-2020-authorship}. The rise of pre-trained LMs has stimulated various improvements of the detectors. The RoBERTa model \citep{liu2019roberta} has demonstrated an outstanding performance with respect to many TGMs and domains \citep{adelani2020generating,fagni2020tweepfake}. The capabilities of generative models such as GROVER \citep{zellers2019defending} and GPT-2 \citep{radford2019language} have been also evaluated on the task \citep{bahri2020generative}. Last but not least, \citet{bakhtin2019real} discriminate artificial texts by training a ranking energy-based model over the outputs of a pre-trained LM.

\paragraph{Unsupervised Artificial Text Detectors} Another line of methods incorporates probability-based measures combined with a set of pre-defined thresholds \citep{solaiman2019release}. Such methods open up a possibility of the \emph{human in the loop} approach where a human makes decisions with the help of pre-trained LMs \citep{ippolito-etal-2020-automatic}. The GLTR tool \citep{gehrmann-etal-2019-gltr} supports human-model interaction by visualizing the properties of a text inferred by the model, which improves the human detection rate of artificial texts. A promising direction is involving acceptability and pseudo-perplexity metrics \citep{lau-etal-2020-furiously,salazar-etal-2020-masked} that can be used to evaluate text plausibility.

\section{Background}
\subsection{BERT Model}
BERT is a transformer-based LM that has pushed state-of-the-art results in many NLP tasks. The BERT architecture comprises $L$ encoder layers with $H$ attention heads in each layer. The input of each attention head is a matrix $X$ consisting of the $d$-dimensional representations (row-wise) of $m$ tokens, so that $X$ is of shape $m \times d$. The head outputs an updated representation matrix  $X^\mathrm{out}$:

\begin{equation}
\begin{split}
X^\mathrm{out} & = W^\mathrm{attn}(XW^{\mathrm V}) \\
 \mbox{ with } 
 W^\mathrm{attn} & = \mathrm{softmax}\left(\frac{(XW^{\mathrm Q})(XW^\mathrm{K})^\mathrm{T}}{\sqrt{d}}\right),
\label{eqn:attention}
\end{split}
\end{equation}
where $W^\mathrm{Q}$, $W^\mathrm{K}$, $W^\mathrm{V}$ are trained projection matrices of shape $d \times d$ and $W^\mathrm{attn}$ is of shape $m \times m$  matrix of attention weights. Each element $w^\mathrm{attn}_{ij}$ can be interpreted as a weight of the $j$-th input's \emph{relation} to the $i$-th output: larger weights mean stronger connection between the two tokens. 

\subsection{Attention Map and Attention Graph}
An {\it attention map} displays an attention matrix $W^{attn}$ (Equation \ref{eqn:attention}) in form of a heat map, where  the color of the cell $(i,j)$ represents the \emph{relation} of the $i$-th token to the output representation of the $j$-th token. We use a {\it graph representation} of the attention matrix. The attention matrix is considered to be a weighted graph with the vertices representing tokens and the edges connecting pairs of tokens with strong enough mutual relation (the higher the weight, the stronger the relation). The construction of such graph appears to be quite problematic: a threshold needs to be set to distinguish between weak and strong relations. This leads to instability of the graph's structure: changing the threshold affects the graph properties such as the number of edges, connected components, cycles. The choice of the optimal thresholds is essential to define which edges remain in the graph. TDA methods allow extracting the overall graph's properties which describe the development of the graph with respect to changes in the threshold.

\subsection{Topological Data Analysis} 
TDA instruments permit tracking the changes of a topological structure across varying thresholds for different objects: scalar functions, point clouds, and weighted graph~\cite{chazal2017introduction}. Given a set of tokens $V$ and an attention matrix of pair-wise weights $W$, we build a family of graphs termed as {\it filtration}: an ordered set of graphs for the sequence of increasing thresholds. Figure \ref{fig:M1} depicts the filtration for a toy example. First, we build a graph for a small threshold, using which we filter out the edges with the weights lower than this threshold. Next, we increase the threshold and construct the next graph. Then we compute the core topological features of different dimensions: for $d=0$ these are connected components, for $d=1$ -- ``loops'' (loosely speaking, they corresponds to basic cycles in a graph), and $d$-dimensional ``holes'' for higher dimensions. The amounts of these features at each dimension $\beta_0, \beta_1,...,\beta_d$  are referred to as {\it Betti numbers} and serve as the main invariants of the objects in topology (see Appendix~\ref{app:betti_numbers} for formal definitions).  While the threshold is increasing and the edges are being filtered, new features may arise. For example, the graph can decay into several connected components. At the same time, the features can also disappear when a cycle is broken. For each feature, we check the moment in the filtration when it appears (i.e., its ``birth'') and when it disappears (i.e., its ``death''). These moments are depicted on a diagram called {\it barcode} (see Figure \ref{fig:M1}). The barcode's horizontal axis corresponds to the sequence of thresholds. Each horizontal line (``bar'') corresponds to a single feature (``hole''): the line lasts from the feature's ``birth'' to the feature's ``death''. Barcodes characterize the ``persistent'' topological properties of the graph, showing how stable topological features are.

\newcommand{\biggestweight}{black!35!green}
\newcommand{\bigweight}{black!45!green}
\newcommand{\aboveaverageweight}{black!70!green}
\newcommand{\mediumweight}{black!25!cyan!10!blue}
\newcommand{\smallweight}{cyan!30!blue!50!}
\newcommand{\smallestweight}{cyan!40!blue!30!white}
\newcommand{\thickedge}{1.8pt}
\begin{figure*}

\begin{tikzpicture}
    \node[inner sep=0pt](pic1) at (0, 0) {
    \includegraphics[scale=0.45]{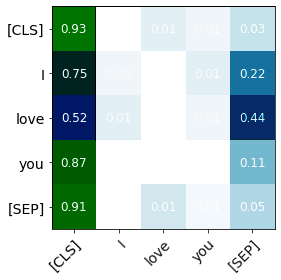}};
    \node[below=0.1cm of pic1] {\Large{\textbf{(a)}}};
    \label{map:i_love_you}
\end{tikzpicture}%
\hspace{0.5cm}
\begin{tikzpicture}[
mydot/.style={
  circle,
  fill,
  inner sep=2pt
},
>=latex,
shorten >= 3pt,
shorten <= 3pt
]
\node[mydot,label={left:[CLS]}] (a1) {}; 
\node[mydot,below=0.7cm of a1,label={left:I}] (a2) {}; 
\node[mydot,below=0.7cm of a2,label={left:love}] (a3) {}; 
\node[mydot,below=0.7cm of a3,label={left:you}] (a4) {};
\node[mydot,below=0.7cm of a4,label={left:[SEP]}] (a5) {};

\node[mydot,right=2cm of a1,label={right:[CLS]}] (b1) {}; 
\node[mydot,below=0.7cm of b1,label={right:I}] (b2) {}; 
\node[mydot,below=0.7cm of b2,label={right:love}] (b3) {}; 
\node[mydot,below=0.7cm of b3,label={right:you}] (b4) {};
\node[mydot,below=0.7cm of b4,label={right:[SEP]}] (b5) {}; 

\path[->] (a1) edge[\biggestweight, line width=\thickedge] (b1);
\path[->] (a2) edge[\aboveaverageweight, line width=\thickedge] (b1)
  edge[\smallweight, line width=\thickedge] (b5);
\path[->] (a3) edge[\mediumweight, line width=\thickedge] (b1)
edge[\mediumweight, line width=\thickedge] (b5);
\path[->] (a4) edge[\bigweight, line width=\thickedge] (b1);
\path[->] (a5) edge[\bigweight, line width=\thickedge] (b1);
\node [below=0.5cm] at (a5)
        {\hspace{1.6cm} {\Large{\textbf{(b)}}}};
\end{tikzpicture}
\hspace{1cm}
\begin{tikzpicture}[
mydot/.style={
  circle,
  fill,
  inner sep=2pt
},
>=latex,
shorten >= 3pt,
shorten <= 3pt
]
\node[mydot,label={[label distance=0.5cm]right:[CLS]}] (a1) {}; 
\node[mydot,below=0.7cm of a1,label={[label distance=0.5cm]right:I}] (a2) {}; 
\node[mydot,below=0.7cm of a2,label={[label distance=0.5cm]right:love}] (a3) {}; 
\node[mydot,below=0.7cm of a3,label={[label distance=0.5cm]right:you}] (a4) {}; 
\node[mydot,below=0.7cm of a4,label={[label distance=0.5cm]right:[SEP]}] (a5) {}; 

\path[->] (a1)edge[\biggestweight, loop left, line width=\thickedge] (a1);
\path[->] (a2)edge[\aboveaverageweight, line width=\thickedge] (a1);
\path[->] (a3)edge[\mediumweight, bend left=15, line width=\thickedge] (a1);
\path[->] (a4)edge[\bigweight, line width=\thickedge, bend left=30] (a1) ;
\path[->] (a5)edge[\bigweight, bend left=45, line width=\thickedge] (a1);

\path[->] (a2)edge[\smallweight, line width=\thickedge, bend left=45] (a5);
\path[->] (a3)edge[\mediumweight, line width=\thickedge, bend left=30] (a5);
\node [below=0.5cm] at (a5)
        {\Large{\textbf{(c)}}};
\end{tikzpicture}
\hspace{0.6cm}

\vspace{1cm}

\begin{tikzpicture}[
mydot/.style={
  circle,
  fill,
  inner sep=2pt
},
>=latex,
shorten >= 3pt,
shorten <= 3pt
]
\node[mydot,label={left:you}] (a1) {}; 
\node[mydot,below=3.44cm of a1,label={left:I}] (a2) {}; 

\node[mydot,below right=1.72cm and 1cm of a1,label={right:[CLS]}] (cls) {};

\node[mydot,right=2cm of a1,label={right:love}] (b1) {}; 
\node[mydot,below=3.44cm of b1,label={right:[SEP]}] (b2) {}; 

\path[->] (a1) edge[\bigweight, line width=\thickedge] (cls);
\path[->] (a2) edge[\aboveaverageweight, line width=\thickedge] (cls)
    edge[\smallweight, line width=\thickedge] (b2);
\path[->] (b1) edge[\mediumweight, line width=\thickedge,] (cls)
    edge[\mediumweight, line width=\thickedge, bend left=30] (b2);
\path[->] (b2) edge[\bigweight, line width=\thickedge] (cls);
\path[->] (cls)edge[loop left, \biggestweight, line width=\thickedge] (cls);
\node [below=0.5cm] at (a5)
        {\hspace{2cm} {\Large{\textbf{(d)}}}};
\end{tikzpicture}
%
\begin{tikzpicture}[
mydot/.style={
  circle,
  fill,
  inner sep=2pt
},
>=latex,
shorten >= 3pt,
shorten <= 3pt
]
\node[mydot,label={left:you}] (a1) {}; 
\node[mydot,below=3.44cm of a1,label={left:I}] (a2) {}; 

\node[mydot,below right=1.72cm and 1cm of a1,label={right:[CLS]}] (cls) {};

\node[mydot,right=2cm of a1,label={right:love}] (b1) {}; 
\node[mydot,below=3.44cm of b1,label={right:[SEP]}] (b2) {}; 

\path[] (a1) edge[\bigweight, line width=\thickedge] (cls);
\path[] (a2) edge[\aboveaverageweight, line width=\thickedge] (cls)
    edge[\smallweight, line width=\thickedge] (b2);
\path[] (b1) edge[\mediumweight, line width=\thickedge,] (cls)
    edge[\mediumweight, line width=\thickedge, bend left=30] (b2);
\path[] (b2) edge[\bigweight, line width=\thickedge] (cls);
\node [below=0.5cm] at (a5)
        {\hspace{2cm} {\Large{\textbf{(e)}}}};
\end{tikzpicture}
\hspace{0cm}
\begin{tikzpicture}[
mydot/.style={
  circle,
  fill,
  inner sep=2pt
},
>=latex,
shorten >= 3pt,
shorten <= 3pt
]
\node[mydot,label={left:you}] (a1) {}; 
\node[mydot,below=3.44cm of a1,label={left:I}] (a2) {}; 

\node[mydot,below right=1.72cm and 1cm of a1,label={right:[CLS]}] (cls) {};

\node[mydot,right=2cm of a1,label={right:love}] (b1) {}; 
\node[mydot,below=3.44cm of b1,label={right:[SEP]}] (b2) {}; 

\path[] (a1) edge[\bigweight, line width=\thickedge] (cls);
\path[] (a2) edge[\aboveaverageweight, line width=\thickedge] (cls);
\path[] (b2) edge[\bigweight, line width=\thickedge] (cls);
\node [below=0.5cm] at (a5)
        {\hspace{2cm} {\Large{\textbf{(f)}}}};
\end{tikzpicture}

\vspace{1cm}

\begin{tikzpicture}[
mydot/.style={
  circle,
  fill,
  inner sep=2pt
},
>=latex,
shorten >= 3pt,
shorten <= 3pt
]
\node[mydot,label={left:you}] (a1) {}; 
\node[mydot,below=3.44cm of a1,label={left:I}] (a2) {}; 

\node[mydot,below right=1.72cm and 1cm of a1,label={right:[CLS]}] (cls) {};

\node[mydot,right=2cm of a1,label={right:love}] (b1) {}; 
\node[mydot,below=3.44cm of b1,label={right:[SEP]}] (b2) {}; 

\node [below=0.5cm] at (a5)
       {\hspace{2cm} {\Large{\textbf{(g)}}}};
\end{tikzpicture}
\hspace{2cm}
\begin{tikzpicture}
    \node[inner sep=0pt](pic2) at (0, 0) {
    \includegraphics[width = 0.6 \textwidth]{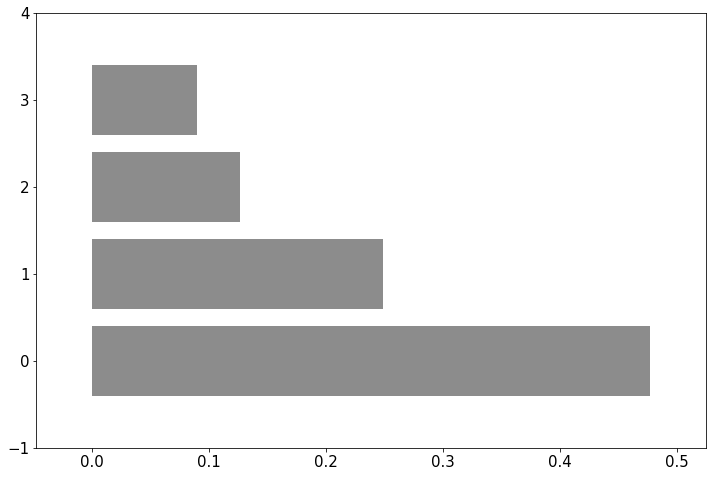}};
    \node[below=0.1cm of pic2] {\Large{\textbf{(h)}}};
  \label{H0pic}
\end{tikzpicture}

\hspace{0.5cm}

\caption{
Let us consider an attention map computed on the sentence ``I love you'' with the BERT model (Layer: 1, Attention Head: 6) which is depicted in~\textbf{(a)}. After matching the vertices of the corresponding graph \textbf{(b)} and removing directions as shown in \textbf{(c)} and \textbf{(d)}, we get graph \textbf{(e)} with 5 vertices and 6 edges. For the sake of better visualization, we do not draw edges with a weight less than 0.2. The graph has one connected component ($\beta_0 = 1$) and two ``loops'' ($\beta_1 = 2$). After filtering out edges with small weight, we get graph \textbf{(f)} which has one new connected component (it is often referred to as ``birth'' of a new component) and does not have any ``loops'' (i.e., the loops that we can see in the previous version of the graph have ``died''). Consequently, after removing all edges, we get graph \textbf{(g)} where $3$ new connected components are born, and now there are $5$ connected components ($\beta_0 = 5$) in total. The barcode \textbf{(h)} depicts $0$-dimensional features (connected components) for the {\it filtration} (\textbf{(e)}, \textbf{(f)} and \textbf{(g)}). Here, the X-axis denotes the filtration parameter $\epsilon$, and the Y-axis denotes the number of the bars. We ignore the ``infinite'' feature persisting through the whole filtration. Note that conventionally on barcodes the $x$ axis is inverted.
} \label{fig:M1}
\end{figure*}

We now detail building the attention graphs, the filtration procedure, and the proposed features which are derived from the attention graphs.

\section{Persistent Features of the Attention Graphs} \label{features}

\paragraph{Informal Definition and Interpretation}
We extract three groups of features from the attention graphs.
{\bf Topological features} (Section \ref{topological_features}) include a set of standard graph properties: the number of connected components, the number of edges, and the number of cycles. These features are calculated for each pre-defined threshold separately and then concatenated. We consider two following variants of the feature calculation: for a directed and an undirected attention graph. {\bf Barcode features} (Section \ref{barcode_features}) are extracted from barcodes.
{\bf Distance to patterns} (Section \ref{distance_features}) is the group of features derived from the attention maps by computing the distance to the attention patterns \citep{clark2019does}. 

\vspace{-2pt}
To give the linguistic interpretation of our features, recall that the graph structures are used in lexicology for describing semantic change laws \citep{hamilton2016diachronic, lipka1990outline, arnold1973english}. The evolution of the meaning of a word with time can be represented as a graph, in which edges represent a semantic shift to different word meanings. Two typical patterns are distinguished in the graph structure: {\it radiation} -- the ``star'' structure, where the primary meaning is connected to other connotations independently; {\it concatenation}, or {\it chaining shift} -- the ``chain'' structure when the connotations are integrated one-by-one. Note that the typical attention patterns \citep{clark2019does} have the same ``radiation'' and ``concatenation'' structure. In pre-trained LMs, the evolution goes through the layers of the model, changing the representation of each token, ending up with highly contextualized token representations, and the aggregated representation of the whole sentence (in the form of the [CLS]-token). 

We consider persistent features as the numerical characteristic of the semantic evolution processes in the attention heads. Topological features deal with clusters of mutual influence of the tokens in the sentence and the local structures like chains and cycles. The barcode features characterize the severity and robustness of the semantic changes. The features with long persistence (large distance between ``birth'' and ``death'') correspond to the stable processes which dominate the others, while short segments in the barcode define processes highly influenced by noise. 
Pattern features provide a straightforward measure of the presence of typical processes over the whole sentence. The so-called ``vertical'' pattern corresponds to the ``radiation'' around the single token when the meaning of the sentence or a part of the sentence is aggregated from all words equally. ``Diagonal'' pattern represents consequent ``concatenation'' structure, going through all the sentence and thus reflecting the dependence of each token's meaning on its left context.

\subsection{Topological Features}
\label{topological_features}

First, we fix a set of thresholds $T = \{t_i\}_{i=1}^{k}, 0 < t_1 < ... < t_k < 1$. Consider an attention head $h$ and corresponding weights $W^\mathrm{attn} = (w^{attn}_{i, j})$. Given a text sample $s$, for each threshold level $t \in T$ we define the weighted directed graph $\Gamma^{h}_s(t)$ with edges $\{j \to i \mid w^\mathrm{attn}_{ij} \ge t\}$ and its undirected variant  $\overline{\Gamma^{h}_s(t)}$ by setting an undirected edge $v_iv_j$ for each pair of vertices $v_i$ and $v_j$ which are connected by an edge in at least one direction in the graph $\Gamma^{h}_s(t)$. 

We consider the following features of the graphs: 
\begin{itemize}[topsep=0pt,itemsep=-1ex,partopsep=1ex,parsep=1ex]
\item the first two Betti numbers of the undirected graph $\overline{\Gamma^{h}_s(t)}$. The feature calculation procedure is described in Appendix \ref{app:top_algorithm};
\item  the number of edges ($\mathbf{e}$), the number of strongly connected components ($\mathbf{s}$) and the amount of simple directed cycles ($\mathbf{c}$) in the directed graph $\Gamma^{h}_s(t)$.
\end{itemize} 

To get the whole set of topological features for the given text sample $s$ and the attention head $h$, we concatenate the features for all the thresholds, starting from~$T$.

\subsection{Features Derived from Barcodes}
\label{barcode_features}
For each text sample we calculate barcodes of the first two persistent homology groups (denoted as $H_0$ and $H_1$) on each attention head of the BERT model (see Appendix~\ref{app:betti_numbers} for further details). We compute the following characteristics of these barcodes:
\begin{itemize}[topsep=0pt,itemsep=-1ex,partopsep=1ex,parsep=1ex]
    \item 
    The sum of lengths of bars;
    \item 
    The mean of lengths of bars;
    \item 
    The variance of lengths of bars;
    \item 
    The number of bars with time of birth/death greater/lower than threshold;
    \item 
    The time of birth/death of the longest bar (excluding infinite);
    \item 
    The overall number of bars;
    \item 
    The entropy of the barcode.
\end{itemize}

\begin{table*}[!ht] 
\centering
\begin{threeparttable}
\resizebox{\textwidth}{!}{
\begin{tabular}{cccccccccccc}
\toprule 
\multicolumn{2}{c}{\textbf{Text Source}} & \multicolumn{2}{c}{\textbf{Train}} & \multicolumn{2}{c}{\textbf{Validation}} & \multicolumn{2}{c}{\textbf{Test}} &
\multicolumn{2}{c}{\textbf{$|$Vocab$|$}} &
\multicolumn{2}{c}{\textbf{Length}} 
\\ \cmidrule{3-12} 

\multicolumn{2}{l}{} & \textbf{H} & \textbf{M} & \textbf{H} & \textbf{M} & \textbf{H} & \textbf{M} & 
\textbf{H} & \textbf{M} & \textbf{H} & \textbf{M} 
\\ \midrule
WebText & \makecell{GPT-2 Small;\\ pure sampling} & 20K & 20K & 2.5K & 2.5K & 2.5K & 2.5K  & 220K & 532K & 593 \tiny{$\pm$ 177} & 515 \tiny{$\pm$ 322} \\ 
\makecell{Amazon \\ Review} & \makecell{GPT-2 XL\\ pure sampling} & 5K & 5K & 1K & 1K & 4K & 4K & 47K & 49K & 179 \tiny{$\pm$ 170} & 177 \tiny{$\pm$ 171} \\
RealNews &  \makecell{GROVER \\ top-$p$ sampling} & 5K & 5K & 1K & 1K & 4K & 4K & 98K & 75K & 721 \tiny{$\pm$ 636} & 519 \tiny{$\pm$ 203} \\
\bottomrule
\end{tabular}
}
\caption{Statistics for the datasets used in the experiments on the artificial text detection task. \textbf{H}=Human; \textbf{M}=Machine.}
\label{tab:dataset_stata}
\end{threeparttable}
\end{table*}

\subsection{Features Based on Distance to Patterns}
\label{distance_features}
The shape of attention graphs in distinct attention heads can be divided into several patterns \cite{clark2019does}. We hypothesize that appearance of such patterns in a particular head or ``intensity'' of the pattern (i.e., the threshold $t$ on which the pattern appears) may carry essential linguistic information. Thus, we formalize these attention patterns and calculate the distances to them as follows.

Let $A = (a_{ij})$ be an incidence matrix of the graph $\Gamma$ with $n$ vertices, where $a_{ij} = 1$ for all edges $(ij) \in E$ and $0$ for all other $i, j$.
Let $\Gamma = (V, E)$ and $\Gamma' = (V, E')$ be two graphs with the same set of vertices, and let $A$, $A'$ be their incidence matrices.
As a distance $d$ between such graphs we use Frobenius norm of the difference 
$|| A - A'||_F = \sqrt{\sum_{i,j} (a_{ij}-a'_{ij})^2},$ normalized by the norms of the matrices of compared graphs:
$$
    d(\Gamma, \Gamma') = \frac{|| A - A'||_F}{\sqrt{||A||_F^2 + ||A'||_F^2}} $$
$$
    = \sqrt{\frac{\sum_{i,j} {(a_{ij}-a'_{ij})}^2}{\sum_{i,j} (a_{ij}^2 + {a'_{ij}}^2)}}.
$$
Such distance takes values between 0 and 1. For the unweighted graphs we have:
$$
d(\Gamma, \Gamma') = \sqrt{\frac{|E \triangle E'|}{|E|+|E'|}},
$$
where $E \triangle E' = (E \backslash E') \bigcup (E' \backslash E)$ is the symmetric difference of sets $E$ and $E'$.

We consider distances from the given graph $\Gamma$ to attention patterns $\Gamma_i$ as the graph features $ d_i (\Gamma ) = d(\Gamma, \Gamma_i)$, and the patterns posed by \citep{clark2019does}:

\begin{itemize}[topsep=0pt,itemsep=-1ex,partopsep=1ex,parsep=1ex]
    \item Attention to the previous token.    $\Gamma_{feature}: E = (i+1,i)$, $i=\overline{1, n-1}$.

    \item Attention to the next token.
    $\Gamma_{feature}: E = (i,i+1)$, $i=\overline{1, n-1}$.

    \item Attention to [CLS]-token.
    [CLS]-token corresponds to the vertex 1 of the set $V = [1,n]$ as it denotes the beginning of the text.
    $ \Gamma_{feature}: E = (i,1)$, $i=\overline{1, n}$.
    
    \item  Attention to [SEP]-token.
    Suppose 
    $i_1,  \dots, i_k$ are the indices of [SEP]-tokens. Then $\Gamma_{feature}: E = (i,i_t)$, $i= \overline{1,n}$, $t = \overline{1,k}$.

    \item Attention to punctuation marks.
    Let $i_1, \dots, i_k$ be the indices of the tokens which correspond to commas and periods. $\Gamma_{feature}: E = (i,i_t)$, 
    $i= \overline{1,n}$, $t = \overline{1,k}$. Note that this pattern can be potentially divided into Attention to commas and Attention to periods.

\end{itemize}

\section{Experiments}
\label{experimentssection}
\subsection{Artificial Text Detection}
\label{detection-task}
\paragraph{Data}
We prepare three datasets from different domains to conduct the experiments on the task of artificial text detection. Table \ref{tab:dataset_stata} outlines statistics for the datasets. Each split is balanced by the number of samples\footnote{Each sample is truncated to 128 BertTokenizer tokens (\texttt{bert-base-uncased}).
} per each target class.

\vspace{0.5em}\noindent \textbf{WebText \& GPT-2} comprises a subset of natural and generated texts from the datasets proposed by \citet{gpt2paper}. (i) \textbf{WebText} contains filtered and de-duplicated natural texts from Reddit; (ii) \textbf{GPT-2 Output Dataset}\footnote{\url{https://github.com/openai/gpt-2-output-dataset}} includes texts generated by various versions of the GPT-2 model fine-tuned on \textbf{WebText}. We use texts generated by GPT-2 Small (117M) with pure sampling. 

\vspace{0.5em}\noindent \textbf{Amazon Reviews \& GPT-2} consists of a subset of Amazon product reviews \citep{amazon_review} and texts generated by GPT-2 XL (1542M) with pure sampling, fine-tuned on this dataset \citep{solaiman2019release}.

\vspace{0.5em}\noindent \textbf{RealNews \& GROVER} ~\citep{zellers2019defending} includes a subset of the news articles from RealNews (that are not present in the GROVER training data) and news articles generated by GROVER with top-$p$ sampling.

\begin{table*}[t!]
\centering
\small
\begin{tabular}{lccc} 
 \toprule 
  \textbf{Model} & \makecell{\textbf{WebText \&} \\ \textbf{GPT-2 Small}} & \makecell{\textbf{Amazon Reviews \&} \\ \textbf{GPT-2 XL}} & \makecell{\textbf{RealNews} \& \\ \textbf{GROVER}} \\ [0ex] 
 \midrule
 \textbf{TF-IDF, N-grams} & 68.1 & 54.2 & 56.9 \\

 \textbf{BERT [CLS trained]} & 77.4 & 54.4 & 53.8 \\

 \textbf{BERT [Fully trained]} & \textbf{88.7} & \textbf{60.1} & \textbf{62.9} \\

 \textbf{BERT [SLOR]} & 78.8 & 59.3 & 53.0 \\
 \midrule

 \textbf{Topological features} & 86.9 & 59.6 & 63.0 \\

 \textbf{Barcode features} & 84.2 & 60.3 & 61.5 \\

 \textbf{Distance to patterns} & 85.4 & 61.0 & 62.3 \\
 \midrule
 \textbf{All features} & \textbf{87.7} & \textbf{61.1} & \textbf{63.6} \\
\bottomrule
\end{tabular}
\smallskip
\caption{The results of the artificial text detection experiments. The performance is measured by the accuracy score~(\%).
}
\label{table_results}
\end{table*}

\paragraph{Baselines}
We use \texttt{bert-base-uncased}\footnote{\url{https://huggingface.co/bert-base-uncased}} model from the HuggingFace library \citep{wolf2019huggingface} for the BERT-based baselines described below. (i)~\textbf{BERT [CLS trained]} is a linear layer trained over [CLS]-pooled text representations. Note that the weights of the BERT model remain frozen. (ii) \textbf{BERT [Fully trained]} is a fully fine-tuned BERT model. We also train Logistic Regression classifier from scikit-learn library \citep{pedregosa2011scikit} over (iii) \textbf{TF-IDF, N-grams} with the N-gram range $\in [1, 2]$ and (iv) \textbf{BERT [SLOR]} \cite{pauls2012large}, an pseudo-perplexity-based acceptability measure inferred with the BERT model under the implementation by \citet{lau-etal-2020-furiously}\footnote{\url{https://github.com/jhlau/acceptability-prediction-in-context}}. 

\paragraph{Models} We train Logistic Regression classifier over the persistent graph features derived from the attention matrices from the BERT model: (i) \textbf{Topological features} (Section \ref{topological_features}), (ii) \textbf{Barcode features} (Section \ref{barcode_features}) and (iii) \textbf{Distance to patterns} (Section \ref{distance_features}). (iv) \textbf{All features} is the concatenation of the features mentioned above. The training details for the baselines and models are outlined in Appendix \ref{app:train_details}.

\paragraph{Results} Table \ref{table_results} outlines the results of the artificial text detection experiments on the three datasets. Note the diversity of the experiment setting where the methods are tested with respect to the TGM, TGM's size, the decoding method, domain, and stylistic properties (texts from the \textbf{Amazon Reviews \& GPT-2} are shorter as compared to those of \textbf{WebText \& GPT-2} and \textbf{RealNews \& GROVER}). The overall tendency is that the proposed TDA-based classifiers outperform the count-based (\textbf{TF-IDF, N-grams}) and two BERT-based baselines (\textbf{BERT [CLS trained]}, \textbf{BERT [SLOR]}) up to 10\%. The concatenation of the features achieves the performance on par with the fully trained BERT model on all datasets.

\subsection{Robustness towards Unseen Models}
\label{robustness}
This setting tests the robustness of the artificial text detection methods towards unseen TGMs on the \textbf{WebText \& GPT-2} dataset. The baselines and models are trained on texts from the GPT-2 small model and further used to detect texts generated by unseen GPT-style models with pure sampling: GPT-2 Medium (345M), GPT-2 Large (762M) and GPT-2 XL (1542M). Note that such a setting is the most challenging as it requires the transfer from the smallest model to that of the higher number of parameters ~\citep{jawahar2020automatic}.

\begin{figure}[t!]
  \includegraphics[width=\linewidth]{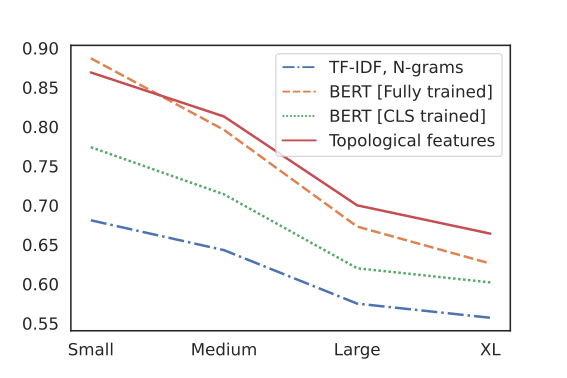}
  \caption{The results of the robustness experiments. X-axis=GPT-2 model size. Y-axis=Accuracy score.}
  \label{quality-drop-plots}
\end{figure}

\paragraph{Results}
Figure \ref{quality-drop-plots} demonstrates the results on the robustness experimental setup. The simple linear classifier trained over the \textbf{Topological features} demonstrates the minor performance drop on the task of detecting artificial texts by the larger GPT-style models as opposed to the considered methods. However, the TDA-based classifier performs slightly worse than \textbf{BERT [Fully trained]} on the test subset by GPT-2 Small.

\subsection{Attention Head-wise Probing}
\label{attention-probing}

\begin{figure*}[htp!]
    \centering
    \begin{subfigure}[b]{0.47\linewidth}
    \centering
    \includegraphics[width=\linewidth]{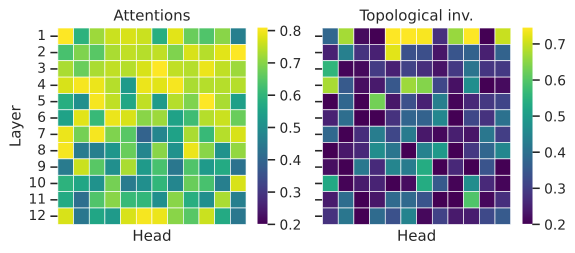}
    \caption{\textsc{Length}}
    \label{fig:heatmap-length}
    \end{subfigure}
    \begin{subfigure}[b]{0.47\linewidth}
    \centering
    \includegraphics[width=\linewidth]{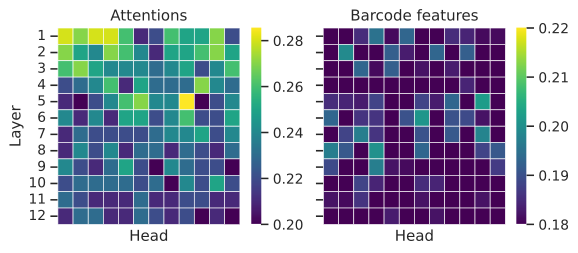}
    \caption{\textsc{Depth}}
    \label{fig:heatmap-depth}
    \end{subfigure}    
    \caption{Heat maps of attention head-wise probing on \textsc{Length} (Left) and \textsc{Depth} (Right) tasks. Attentions=Frozen attention weights.
    X-axis=Head index number. Y-axis=Layer index number.
    The brighter the color, the higher the accuracy score for the attention head.
    }
    \label{fig:probe_heatmaps}
\end{figure*}

\paragraph{Data}
SentEval \citep{conneau2018you} is a common probing suite for exploring how various linguistic properties are encoded in the model representations. The probing tasks are organized by the type of the property: \emph{surface}, \emph{syntactic} and \emph{semantic}. We use the undersampled tasks\footnote{Each probing task is split into 25K/5K/5K train/validation/test sets. The sets are balanced by the number of instances per each target class.} to analyze what properties are stored in the topological features.

\paragraph{Method} Attention head-wise probing \citep{jo2020roles} allows investigating the patterns of how attention heads from each layer of the model contribute most to a probing task. Logistic Regression is trained over the intermediate outputs of the model $h_{i,j}$, where \emph{i} and \emph{j} denote the indices of the layer and the attention head. We use the publicly available code\footnote{\url{https://github.com/heartcored98/transformer_anatomy}} to train the classifier over two groups of the input features: (i) the intermediate outputs $h_{i,j}$ produced by the frozen BERT model and (ii) the topological features derived from $h_{i,j}$ as outlined in Sections \ref{topological_features}, \ref{barcode_features}. The performance is evaluated by the accuracy score, and the heat maps of the probing scores are constructed to introspect how a certain linguistic property is distributed across different layers and attention heads. Refer to \citet{jo2020roles} for more details.

\paragraph{Results} The results demonstrate that the topological features tend to be sensitive to the surface and syntactic properties as opposed to the semantic ones.
Figure \ref{fig:probe_heatmaps} shows heat maps of the attention head-wise evaluation on \textsc{Length} (Figure \ref{fig:heatmap-length}, \emph{surface} property) and \textsc{Depth} (Figure \ref{fig:heatmap-depth}, \emph{syntactic} property) tasks\footnote{\textsc{Length} is a 6-way classification task and \textsc{Depth} comprises 7 classes denoting the depth of a syntax tree.}. While the sentence length is distributed across the majority of the frozen attention heads, specifically at the lower-to-middle layers~$[1-8]$, the topological features capture the property at layer~$[1]$ and by fewer heads at layers~$[2, 4-5, 9-11]$. The depth of the syntax tree is encoded in the frozen heads at the lower-to-middle layers~$[1-5]$, whereas the barcode features predominantly localize the property at the middle-to-higher layers~$[5-9]$. 

The overall pattern for the surface and syntactic tasks is that the persistent graph features can \emph{lose} some information on the linguistic properties during the derivation of the features from the attention matrices. The localization of the properties after the derivation gets changed, and the head-wise probe performance may significantly decrease. Notably, the majority of the semantic tasks receive rapid decreases in the probe performance on the persistent graph features as compared to the frozen heads. The reason is that the features operate purely on the surface and structural information of the attention graph, leaving semantics unattended.

\section{Discussion}

\begin{figure}[t!]
    \includegraphics[width=0.5\textwidth]{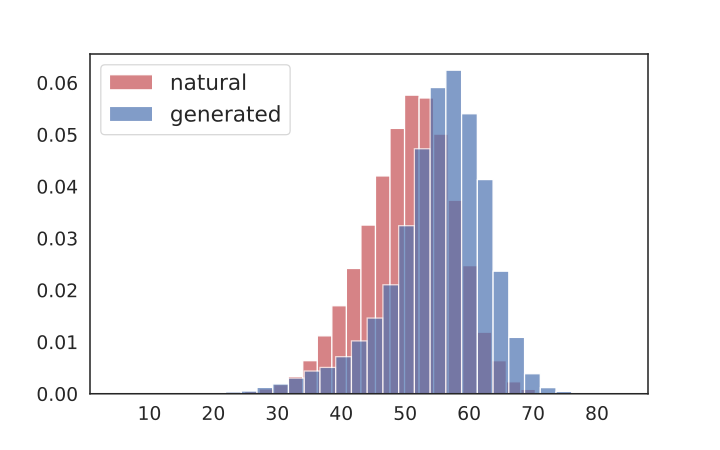}
    \caption{The distribution shift of the sum of the bars in $H_0$ between the natural and generated texts on the \textbf{WebText \& GPT-2 dataset} (Layer: 9; Head: 7). TGM: GPT-2 Small with pure sampling.}
    \label{fig:h0_dist}
\end{figure}

\paragraph{Structural Differences between Natural and Generated texts} The TDA-based classifiers rely on the  structural differences in the topology of the attention maps to distinguish between natural and generated texts. Figure~\ref{fig:h0_dist} shows that the distributions of the sum of bars in $H_0$ differ for natural and generated texts. For the former, it is shifted to the left. We provide more examples of the distribution shift for different heads and layers in Figure~\ref{fig:dist}, Appendix~\ref{app:betti_numbers}. The weights for natural texts are concentrated more on the edges of the maximum spanning tree (MST), so that the model focuses on the sentence structure, or on the  ``skeleton'' of the MST. The weights for the artificially generated texts are distributed more evenly among all edges. As the TDA-based classifiers appear to be robust towards unseen TGMs, we may conclude that such structural properties are inherent to the models of different sizes, so that shifts in the distribution of the sum of bars in $H_0$ hold for texts generated by different TGMs. This feature appears to be the key one as utilizing it alone for the prediction provides us with the 82\% accuracy score on the \textbf{WebText \& GPT-2} dataset.

\paragraph{Semantics is Limited} 
The TDA-based methods do not take the semantic word similarity into account, as they only capture inter-word relations derived from the attention graphs. The probing analysis supports the fact that the features do not encode the semantic properties, carrying only surface and structural information. However, this information appears to be sufficient for the considered task.

\paragraph{Time Complexity}
The attention matrices are computed each time when an input sample is fed to BERT. It follows that the computational complexity of our methods can not be lower than the one for BERT’s complexity itself, which makes asymptotically $O(n^2d + nd^2)$ per one attention head \cite{vaswani2017attention}, where $n$ is the sequence length, and $d$ is the words embedding dimension. On the other hand, the calculation of the topological features by thresholds (given that the number of thresholds is constant), aside of the number of simple cycles, features of $0$-dimensional barcodes, and features based on the distance to patterns are linear by the number of edges of the attention graphs. This means that for at least these features we do not go beyond the asymptotic complexity of the BERT model inference, even for sparse attention variants.

The number of simple cycles and the features of $1$-dimensional barcodes are more computationally expensive. Note that omitting these features provides a significant speed up with minor performance drops. 


\section{Conclusion}
This paper introduces a novel method for the task of artificial text detection based on TDA. We propose three types of interpretable topological features that can be derived from the attention maps of any transformer-based LM. The experiments demonstrate that simple linear classifiers trained on these features can outperform count- and neural-based baselines, and perform on par with a fully fine-tuned BERT model on three common datasets across various domains. The experimental setup also highlights the applicability of the features towards the TGM architecture, TGM's size and the decoding method. Notably, the TDA-based classifiers tend to be more robust towards unseen GPT-style TGMs as opposed to the considered baseline detectors. The probing analysis shows that the features capture surface and structural properties, lacking the semantic information. A fruitful direction for future work is to combine the topological features with those that encode the semantics of the input texts, and test the methods on a more diverse set of the TGM architectures, decoding methods and transformer LMs to infer the attention graphs from. We are publicly releasing the code, hoping to stimulate the research on the TDA-based investigation of the inner workings of the transformer-based models and the applicability of TDA methods to other NLP tasks.

\section*{Acknowledgement}
The work of Serguei Barannikov and Evgeny Burnaev, related to topological data analysis and machine learning in Sections \ref{features} and \ref{experimentssection}, is supported by Ministry of Science and Higher Education grant No. 075-10-2021-068. Ekaterina Artemova is supported by the framework of the HSE University Basic Research Program.


\bibliography{anthology,custom}
\bibliographystyle{acl_natbib}

\appendix
\newpage
\clearpage



\section*{Appendices}
\section{Topological Features Calculation}
\label{app:top_algorithm}
\begin{algorithm}
	\caption{Topological Features Calculation} 
    \label{our_alg}
	\begin{algorithmic}[2]
    	\smallskip
	    \Require {Text sample $s$}
	    \Require {Set of chosen attention heads $H_M$ of model $M$}
	    \Require {Thresholds array $T$}
	    \Require{Topological feature $f$ of unweighted graph}
	    \smallskip
	    \Ensure {Features array $Features$}
		\medskip
		\Procedure{features\_calc}{$s, H_M, T$}
	        \ForAll {$h \in H_M, t \in T$}
		        \State Calculate attention graph $\mathsf{\Gamma^{h}_s} = (V, E, W^{att}_{h, s})$ on sample $s$ on head $h$
		            \State $E^h_s(t) \leftarrow \{e \in E(\mathsf{\Gamma^{h}_s}):$ $ W^{att}_{h, s}(e) \geq t\}$ 
    				\State $\Gamma^{h}_s(t) \leftarrow (V, E^h_s(t))$
    				\State $\overline{E^h_s(t)} \leftarrow \left\{ \{i, j\} : (i, j) \in E^h_s(t) \right\} $
    				\State $\overline{\Gamma^h_s(t)} \leftarrow (V, \overline{E^h_s(t))}$ 
    				\State Calculate 
    				$f(\overline{\Gamma^h_s(t)})$ 
    				
    			\EndFor
			\State $Features \leftarrow \left[ f(\overline{\Gamma^h_s(t)})\right]_{t \in T}^{h \in H_M}$ 
    		\State \textbf{return} $Features$ 
    	\EndProcedure
	\end{algorithmic} 
\end{algorithm}

\begin{figure*}[htb!]
  \includegraphics[width=\linewidth]{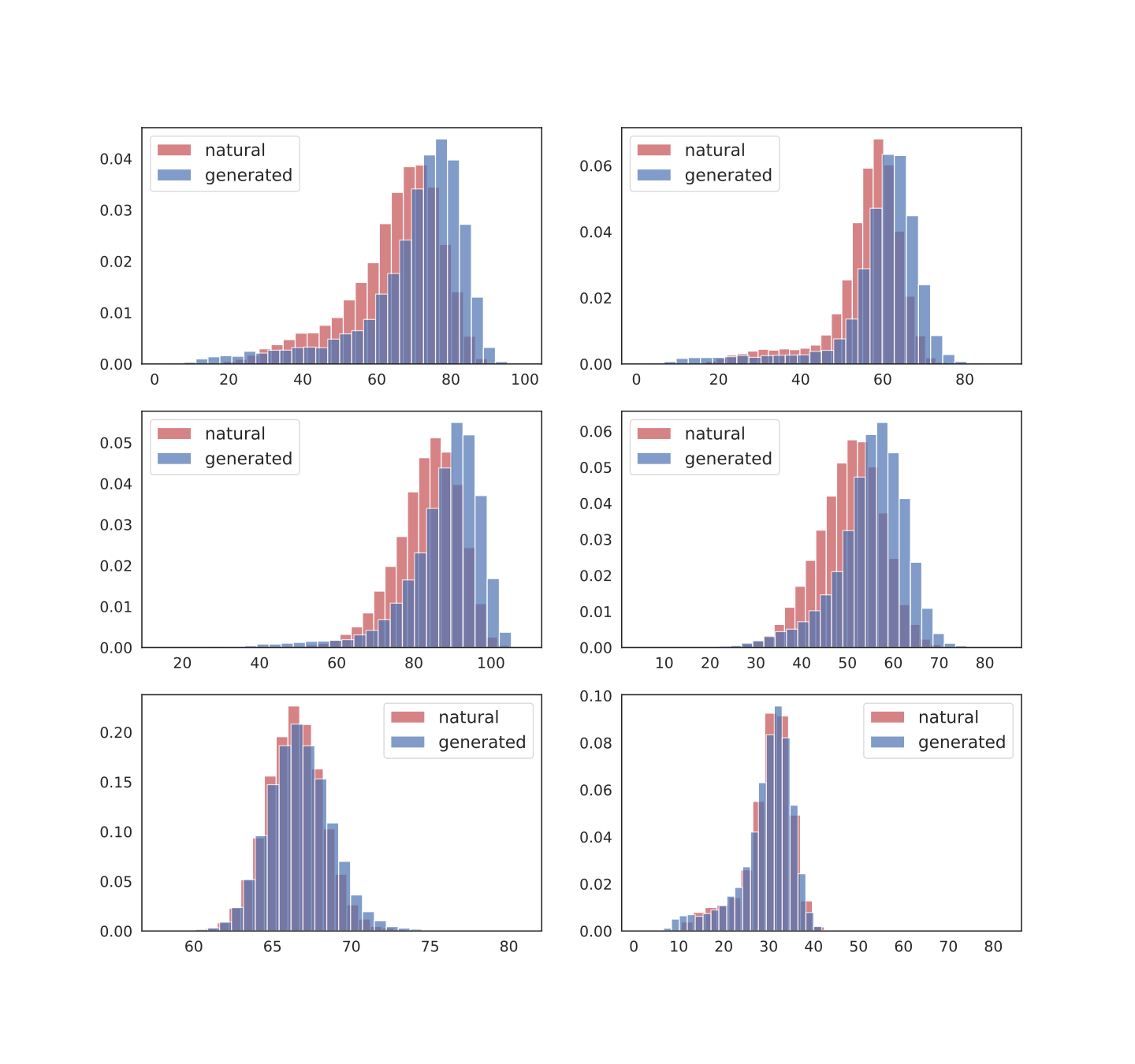}
      \caption{The distribution shift of the sum of the bars in $H_0$ between the natural and generated texts 
      for different layers and attention heads on the \textbf{WebText \& GPT-2} dataset. Top=(Layer: 7; Head: 5), (Layer: 8; Head: 10); Middle=(Layer: 9; Head: 11), (Layer: 9; Head: 7); Bottom=(Layer: 0; Head: 3), (Layer: 5; Head: 6). TGM: GPT-2 Small with pure sampling.}
  \label{fig:dist}
\end{figure*}

\section{Persistent Homology and Betti Numbers}
\label{app:betti_numbers}
Recall that a simplicial complex $K$ is a collection of subsets of a finite set called {\em simplices} such that each subset of any element of $K$ also is an element of $K$; such subsets of a simplex are called {\em faces}. 
In particular, an undirected graph is a simplicial complex where the simplices correspond to the edges and vertices of the graph.  
The set of all formal $\mathbb{Z}$-linear combinations of the $p$-dimensional  simplices (that is, $(p+1)$-element finite sets from the collection) of $K$  is denoted $\mathsf{C}_{p}(K)$.  These linear combinations $c =\sum_{j} \gamma_{j}\sigma_{j}$ are called {\em $p$-chains}, where the $\gamma_j \in \mathbb{Z}$ and the $\sigma_{j}$ are $p$-simplices in $K$.

The boundary, $\partial(\sigma_j)$, is the formal sum of the $(p-1)$-dimensional faces of $\sigma_j$ and the boundary of the chain is obtained by extending $\partial$ linearly,
\begin{equation}\nonumber
\partial(c) = \sum_{j} \gamma_j\partial(\sigma_j)
\end{equation}
for $c$ as above.

The $p$-chains that have boundary $0$ are called $p$-{\em cycles}, they form a subgroup
$\mathsf{Z}_{p}(K)$ of $\mathsf{C}_{p}(K)$. 
The $p$-chains that are the boundary of $(p + 1)$-chains are called $p$-boundaries and form a subgroup $\mathsf{B}_{p}(K)$ of $\mathsf{C}_{p}(K)$.
The quotient group $\mathsf{H}_p(K) = \mathsf{C}_{p}(K) / \mathsf{B}_{p}(K)$ is called the $p$-th {\em homology} of $K$. 
Their ranks $\beta_p = rk (\mathsf{H}_p(K))$ of these abelian groups are called {\em Betti numbers}. The homology and the Betti numbers 
are classical topological invariants of $K$.

In particular, a graph $G = (E,V)$ contains  0-dimensional and 1-dimensional faces. It follows that its topological form is essentially described by the numbers $\beta_0$ and $\beta_1$, which are the only nonzero Betti numbers. Here $\beta_0$ is the number of connected components of $G$, and $\beta_1$ is the number of independent cycles of the graph (which is equal to $|E|-|V|+\beta_0$).

A {\em subcomplex} of $K$ is a subset of simplices that is closed under the face relation. A {\em filtration} of $K$ is a nested sequence of subcomplexes that starts with the empty complex and ends with the complete complex,
\begin{equation}\nonumber
\emptyset  \subset K_{t_1} \subset K_{t_2}   \subset \cdots \subset K_{t_m} = K.
\end{equation} 
In particular, to any weighted undirected graph
$G = (V,E)$ 
one can associate naturally the filtration 
\begin{equation}
\label{eq:graph_filtration}
\emptyset   \subset G_{t_1} \subset \dots \subset G_{t_m} = G,
\end{equation}
where $\{t_i\}$ is the set of all weights of the edges, $G_{t_i} = (V,E_{t_i})$ and $E_{t_i}$ consists of all edges of $E$ with weight less or equal to $t_i$.

In our calculations, the increasing filtration is obtained by reversing the attention matrix weights: $w\mapsto 1-w$.

The $p$-th persistent
homology of $K$ is the pair of sets of vector spaces $\{
\mathsf{H}_p(K_{t_i}) | 0\le i\le l \}$
and maps
$\{ 
f_{i,j}: \mathsf{H}_p(K_{t_i}) \to 
\mathsf{H}_p(K_{t_j})| 1\le i < j \le l
\}$, where the maps are 
induced by the inclusion maps $K_{t_i}\to K_{t_j}$.

Each persistent homology class $\alpha$ in this sequence is ``born'' at some $K_{t_i}$ and ``dies'' at some $K_{t_j}$ or never dies \citep{barannikov2021canonical, B94}. One can visualize this as an interval $[t_i,t_j]$ or $[t_i,+\infty[$. The collection of all such intervals is called the {\em barcode} of the filtration. It is the most useful invariant of the filtration. Note that the information about the persistent homology classes is generally essential to calculate the barcode, whereas the information about the Betti numbers only is insufficient.

In the case of the filtration associated to a weighted graph~(\ref{eq:graph_filtration}),
the $H_0-$th barcode (respectively, $H_1-$th) consists of the intervals of the form $[0, d_i]$ (resp., $[s_i, \infty[$) only. 
Given a number $l$, the number of intervals of length at most $l$ for $H_0$
(respectively, the number of intervals with the left endpoint at most $l$) is therefore equal to the 
the Betti number $\beta_0 (K_l)$ (resp., $\beta_1 (K_l)$). We see that in this case, we can recover the barcode from the collection of all Betti numbers $\beta_i (K_{t_j})$  where the thresholds set $\{t_j\}$ is the set of all weights of the edges in the graph. On the other hand, a calculation of the $H_0-$barcode gives all Betti numbers $\beta_i (K_{t_j})$ simultaneously for all thresholds.


In TDA, the following filtered simplicial complex is also commonly associated with a graph. The $n$-dimensional simplices of 
the {\em clique complex} (or Whitney complex) $X = X (G)$ of a graph $G = (E,V)$  are the $(n+1)$-cliques of $G$, that is, complete subgraphs with $n+1$ vertices. For example, its $0$-simplices are the vertices, the 1-simplices are the edges, and the 2-simplices are the triangles. The clique complex has a richer topological structure than the graph itself since it may have nonzero Betti numbers $\beta_n (X)$ for $n\ge 2$. 
Note that the $H_0-$barcode of the filtered clique complex coincides with the $H_0-$barcode of ~(\ref{eq:graph_filtration}).
Several software packages are aimed at calculating
the persistent homology of graph clique complex. For this purpose, we have used  Ripser++~\cite{bauer2019ripser,zhang2020gpu}.

The entropy of the barcode is a measure of the entropy of the points in a persistence diagram. Precisely, if we have a barcode as a list of pairs of ``birth'', ``death'': $D = \{(b_i, d_i)\}_{i \in I}$, then the entropy is defined as: $$E(D) = - \sum_{i \in I} p_i \log (p_i)$$ where $p_i = \frac{d_i - b_i}{L_D}$, and $L_D = \sum\limits_{i \in I} (d_i - b_i)$.

Each bar in $H_0-$barcode has the form $[0, d_i]$ where $1-d_i$ is the attention weight of the edge which ``kills'' the  connected component corresponding to this bar.
The sum of lengths of bars in the $H_0-$barcode coincides  with $128-M$ where $M$ is the sum of edge weights of the attention graph maximal spanning tree.  Examples of the shift of distributions of the sum of bars' lengths between natural and generated texts are shown in the top and middle rows in Figure~\ref{fig:dist}.

\section{Training Details}
\label{app:train_details}
\paragraph{Topological Features and TF-IDF} Similar to \citep{solaiman2019release}, the training of the linear classifiers is run with the regularization parameter $L^2$ $\in$ $[1\emph{e}^{-5}, 5\emph{e}^{-5}, \dots, 0.1, 0.5, 1]$ and the maximum number of iterations $max_{iter} \in [1, 2, 3, 5, 10, 100]$ tuned on the validation set. The topological features are concatenated for each attention head for each encoder layer, and further concatenated. The total number of features for each method is equal to $12 \times 12 \times \text{number of features per method}$.

\paragraph{BERT-based} classifiers are trained  with the linear scheduler with the initial learning rate $lr \in [1e^{-5}, 5e^{-5}, 1e^{-4}, \dots , 1e^{-1}, 1]$ and epochs number $e \in [2, 3, 5, 10, 15, 20]$. \textbf{BERT [CLS trained]} is trained for 20 epochs, and \textbf{BERT [Fully trained]} is trained for 2-5 epochs depending on the dataset. We use early stopping controlled by the accuracy on the validation set for each text detection dataset.

\end{document}